# The pitfalls of using open data to develop deep learning solutions for COVID-19 detection in chest X-rays

Rachael Harkness[a,b], Geoff Hall[b,c,d], Alejandro F Frangi[a,b,e,f], Nishant Ravikumar[a,b,*], Kieran Zucker[b,c,d,*]

[a]CISTIB Centre for Computational Imaging and Simulation Technologies in Biomedicine, School of Computing
[b]University of Leeds, Leeds, LS2 9JT, United Kingdom
[c]Leeds Institute of Medical Research at St James's, United Kingdom
[d]Leeds Teaching Hospitals NHS Trust, Leeds, LS9 7TF, United Kingdom
[e]LICAMM Leeds Institute for Cardiovascular and Metabolic Medicine, School of Medicine
[f]LIDA Leeds Institute of Data Analytics
*indicates joint last authorship

## Abstract

*Since the emergence of COVID-19, deep learning models have been developed to identify COVID-19 from chest X-rays. With little to no direct access to hospital data, the AI community relies heavily on public data comprising numerous data sources. Model performance results have been exceptional when training and testing on open-source data, surpassing the reported capabilities of AI in pneumonia-detection prior to the COVID-19 outbreak. In this study impactful models are trained on a widely used open-source data and tested on an external test set and a hospital dataset, for the task of classifying chest X-rays into one of three classes: COVID-19, non-COVID pneumonia and no-pneumonia. Classification performance of the models investigated is evaluated through ROC curves, confusion matrices and standard classification metrics. Explainability modules are implemented to explore the image features most important to classification. Data analysis and model evaluations show that the popular open-source dataset COVIDx is not representative of the real clinical problem and that results from testing on this are inflated. Dependence on open-source data can leave models vulnerable to bias and confounding variables, requiring careful analysis to develop clinically useful/viable AI tools for COVID-19 detection in chest X-rays.*

## Introduction

In line with urgent public interest and clinical need, the field of medical AI research has taken interest in the development of automated solutions that enhance analysis and interpretation of COVID-19-related medical data, with particular focus on the detection of COVID-19 in chest X-rays. A number of deep learning models have been developed with the intention of identifying COVID-19 specific radiological features to alleviate COVID-19 testing bottlenecks [1-3]. Multiple publications and preprints report exceptional model performance, dramatically improving on pneumonia-detection model results from the pre-COVID era and exceeding self-reported radiologist performance [1-3]. Considering the limited diagnostic yield associated with chest X-rays and the challenging nature of distinguishing COVID-19 from other similarly presenting pathologies, these results suggest models may be relying on biases in the training data rather than clinically relevant features.

Due to limited availability of COVID-related imaging data, the vast majority of existing AI methods rely on a heterogeneous mix of open-source data repositories, sourcing non-COVID-19 chest X-rays from larger pre-existing repositories and obtaining COVID-19 chest X-rays from recently released public datasets curated in response to the sudden demand for COVID-related data. Four COVID-19 chest X-ray repositories are used most frequently: (1) COVID-19 Image Data Collection (Cohen) [4], (2) COVID-19 Chest X-ray Dataset Initiative, (3) ActualMed COVID-19 Chest X-ray Dataset Initiative (ActMed) and (4) COVID-19 Radiography Database (SIRM). These four data sources are commonly combined with established pre-covid datasets that comprise chest X-rays of patients with various lung pathologies. One such example is the Radiological Society of North America (RSNA) Pneumonia Detection Challenge dataset [5]. These five data repositories have been collated into one large open-source dataset, termed COVIDx, that has been made readily accessible by its curators [1]. The extensive use of a single public dataset within the community, encourages competition and drives research. However, widespread usage may have fostered an attitude of acceptance and a lack of critical appraisal in the research community, with any dataset errors propagating throughout the research domain without revision. Deep learning models for the detection of COVID-19 from chest X-rays have significant potential for clinical impact, supporting radiologists and acting as a second check against false negative RT-PCR results. The reliability and robustness of these models require rigorous evaluation prior to clinical deployment but due to limited availability of hospital data, existing models have gone largely without such validation. This study aims to investigate the impact of using open-source COVID-19 data by evaluating the performance of three deep learning models proposed in recent publications on both publicly available test data and real-world hospital data. To assess concerns of bias and confounding we also implement explainability modules in the chosen models to show features of significance.

# Methods

## The Data

### Training data

COVIDx was used as the open-source training dataset as it is the largest and most widely used within the community [1]. It contains chest X-rays assigned to three classes; normal, pneumonia and COVID-19. To approximate the data used in previous research the COVIDx dataset was formed using the files made available at : https://github.com/lindawangg/COVID-Net [1]. COVIDx has been updated to include the RSNA International COVID-19 Open Radiology Database (RICORD) dataset. However, this was subsequent to the publication of the models of interest and thus data from this source was excluded in this study [6]. To eliminate the added complications of training on a class imbalanced dataset, cases from the pneumonia and normal class were excluded at random from the COVIDx training set. Figure 1 shows the distribution of chest X-rays included in the COVIDx dataset according to source and class. Overall the balanced COVIDx data contains 4,638 'normal' cases, 4,347 'pneumonia' cases and 3,027 COVID-19 cases. Prior to model training and evaluation, the data sources contributing to the COVIDx dataset were critically assesed with a view to clarify data provenance and identify potential sources of bias. The data, files and code for this analysis are available online: https://bitbucket.org/rkharkness/open-data-study

### External test data

A bespoke test set was created from the CheXpert dataset and the reserved RICORD dataset for external evaluation of the trained models.

1. RICORD

The RICORD dataset provides 1096 COVID-19 chest X-rays from 361 patients and is sourced from four international sites [6].

2. CheXpert

CheXpert is a large public dataset made up of 224,316 chest X-rays from 65,240 patients . The CheXpert labeller, a natural language processing tool, is applied to radiology reports to derive 14 possible labels [7].

Sampling from the CheXpert dataset provides 998 pneumonia-negative and 997 non-COVID-19 pneumonia cases. Class labels were reviewed by a clinical expert to ensure they were clinically appropriate as the pneumonia-negative class includes a wide variety of other lung pathologies, while the non-COVID-19 pneumonia class contains exclusively pneumonia cases (although these may include comorbidities).

### Leeds Teaching Hospital NHS Trust (LTHT) Data

LTHT, a large teaching hospital based in Leeds, UK, provided a dataset of chest X-ray images of patients alongside PCR test results for COVID-19 diagnosis and non-COVID pneumonia diagnosis. A sample of chest X-rays was randomly selected from the LTHT data, supplying 611 COVID-19 cases, 459 pneumonia-negative cases and 299 non-COVID pneumonia cases.

## COVIDx Data Analysis

The provenance and origins of each COVIDx data source was appraised [1]. To evaluate the extent of the misuse of labels observed in the RSNA dataset in the context of model performance labels from RSNA cases in the balanced COVIDx training data were mapped to the original labels associated with the chest X-rays prior to extraction from the larger National Institute of Health (NIH) chest X-ray8 dataset [8]. The proportions of misaligned labels in the pneumonia class of COVIDx were identified.

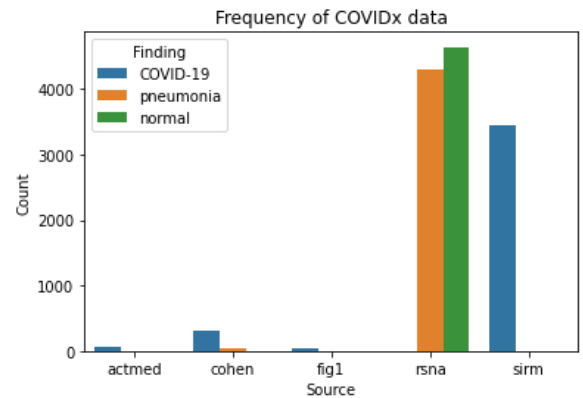

*Figure 1: Frequency of chest X-rays according to class and source.*

## The models

Models were selected based on publication impact factor from those with open-source data and code availability. From these criteria we selected three models: COVID-Net , DarkCovidNet and CoroNet [1-3]. All three models utilise convolutional neural networks (CNNs) for feature learning. While COVID-Net and DarkCovidNet are standard deep CNNs, CoroNet relies on a two stage process to classify images, comprising convolutional autoencoders in the first stage and a standard CNN classifier in the second stage. We have omitted further details regarding these methods for brevity.

### Model training

Models were trained with 3-fold cross validation and hyperparameters were selected according to their described training protocol. Where possible a simple Grad-CAM explainability module was implemented to identify the features that have the greatest impact on classification. Post-training, three rounds of evaluations were undertaken based on (1) COVIDx test dataset (2) The external test dataset and (3) the LTHT data.

### Model evaluation

All models trained on COVIDx training data were evaluated on the COVIDx test data, the external test data and the LTHT data. Weights from each set of cross validation were evaluated. Confusion matrices and ROC curves were generated, and performance metrics were recorded. For each model, the best was selected according to F1 score, ROC curves and confusion matrices were evaluated based from this selection .

# Results

## COVIDx data analysis

Analysis of the balanced COVIDx dataset confirmed misuse of data from the RSNA repository [5]. The RSNA repository, which uses publicly available chest X-ray data from NIH Chestx-ray8 [8], was designed for a segmentation task and as such contains three classes of images, 'Lung Opacity', 'No Lung Opacity/Not Normal', and 'Normal', with bounding boxes available for 'Lung Opacity' cases. In its compilation into COVIDx all chest X-rays from the 'Lung Opacity' class are included in the pneumonia class.

In this task the definition of pneumonia is expanded to include all pneumonia-like lung opacities, as a result, the pneumonia class within the COVIDx dataset contains chest X-rays with an assortment of many other pathologies, including, pleural effusion, infiltration, consolidation, emphysema and masses. Consolidation is a radiological feature of possible pneumonia, not a clinical diagnosis. To use consolidation as a substitute for pneumonia without documenting this is potentially misleading. Of the 4,305 pneumonia cases sourced from the RSNA pneumonia challenge dataset, only 264 were originally labelled as pneumonia, meaning only 6.13% of RSNA pneumonia were accurately labelled within the COVIDx data. Figure 2 shows the frequency of the eight most commonly observed alternative pathologies included in the COVIDx pneumonia data, with all of these pathologies occurring more frequently than pneumonia. Many of these exist as comorbidities in a single chest X-ray, further obscuring the true pneumonia class.

The 'normal' class of COVIDx takes only from the 'Normal' class of the RSNA challenge dataset, meaning all the alternative pathologies contained in the 'No Lung Opacity/Not Normal' class are excluded. While this is in keeping with what is expected within the 'normal' label, expanding the pneumonia class and using only 'normal' chest X-rays, rather than pneumonia-negative cases greatly simplifies the classification task. The end result of this is dataset that reflects a task that is removed from the true clinical problem.

Potential sources of bias are identified in the COVIDx data set. Within its non-COVID data, SIRM blends paediatric chest X-ray images with adult images and is the only significant source of paediatric images within COVIDx. Manual inspection of the ActMed repo revealed the consistent presence of disk-shaped markers in COVID-19 chest X-rays. The collection of heterogenous datasets into COVIDx results in the use of images with a large variation in image sizes. For example, the images from the RSNA dataset are 1024x1024 in resolution, while all SIRM-provided images are a resolution of 299x299. Most models resize images to resolutions between these. To facilitate this, smaller images must be up-sampled and larger images must be down-sampled. This risks generation of artefacts that may bias the model.

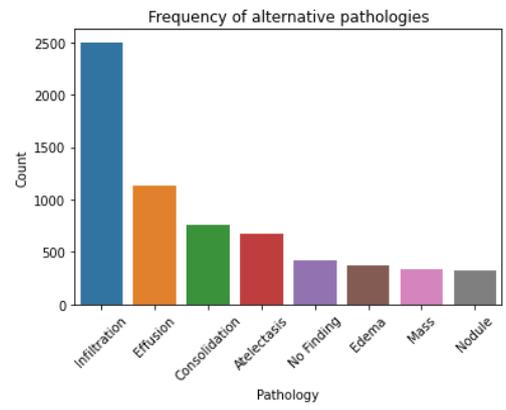

*Figure 2: Counts of the eight most commonly observed pathologies within the COVIDx pneumonia class.*

**Model evaluation**

ROC curves (Figure 3) reflect exceptional performance across all chosen models when tested on the COVIDx test data, with all models achieving area under curves of greater than 0.93 across all classes, exceeding 86% accuracy and reliably separating the COVIDx pneumonia class from the normal class. Testing on non-COVIDx data shows a steep drop in all model performances, this is highlighted by the confusion matrices displayed in Figure 4.

*Table 1– DarkCovidNet test data performance metrics.*

| Test set | Precision | Recall | F1 score | Accuracy |
|---|---|---|---|---|
| COVIDx | 0.87±0.00 | 0.80±0.00 | 0.82±0.00 | 0.88±0.00 |
| External | 0.44±0.00 | 0.43±0.00 | 0.41±0.00 | 0.43±0.00 |
| LTHT | 0.47±0.01 | 0.46±0.00 | 0.44±0.01 | 0.45±0.00 |

*Table 2 - CoroNet test data performance metrics.*

| Test set | Precision | Recall | F1 score | Accuracy |
|---|---|---|---|---|
| COVIDx | 0.81±0.05 | 0.90±0.01 | 0.84±0.05 | 0.88±0.03 |
| External | 0.18±0.07 | 0.34±0.02 | 0.19±0.03 | 0.35±0.01 |
| LTHT | 0.24±0.01 | 0.30±0.00 | 0.15±0.01 | 0.22±0.00 |

*Table 3 – COVIDNet test data performance metrics.*

| Test set | Precision | Recall | F1 score | Accuracy |
|---|---|---|---|---|
| COVIDx | 0.86±0.03 | 0.69±0.05 | 0.72±0.05 | 0.86±0.02 |
| External | 0.34±0.05 | 0.36±0.01 | 0.29±0.02 | 0.38±0.01 |
| LTHT | 0.43±0.01 | 0.39±0.00 | 0.37±0.01 | 0.44±0.03 |

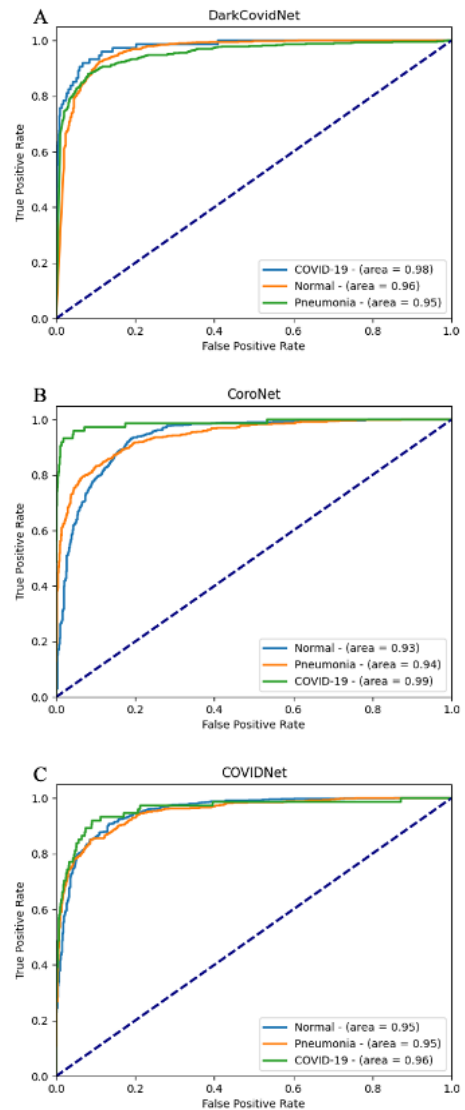

*Figure 3: ROC curves of A) DarkCovidNet B) CoroNet and C) COVIDNet when test on COVIDx test data.*

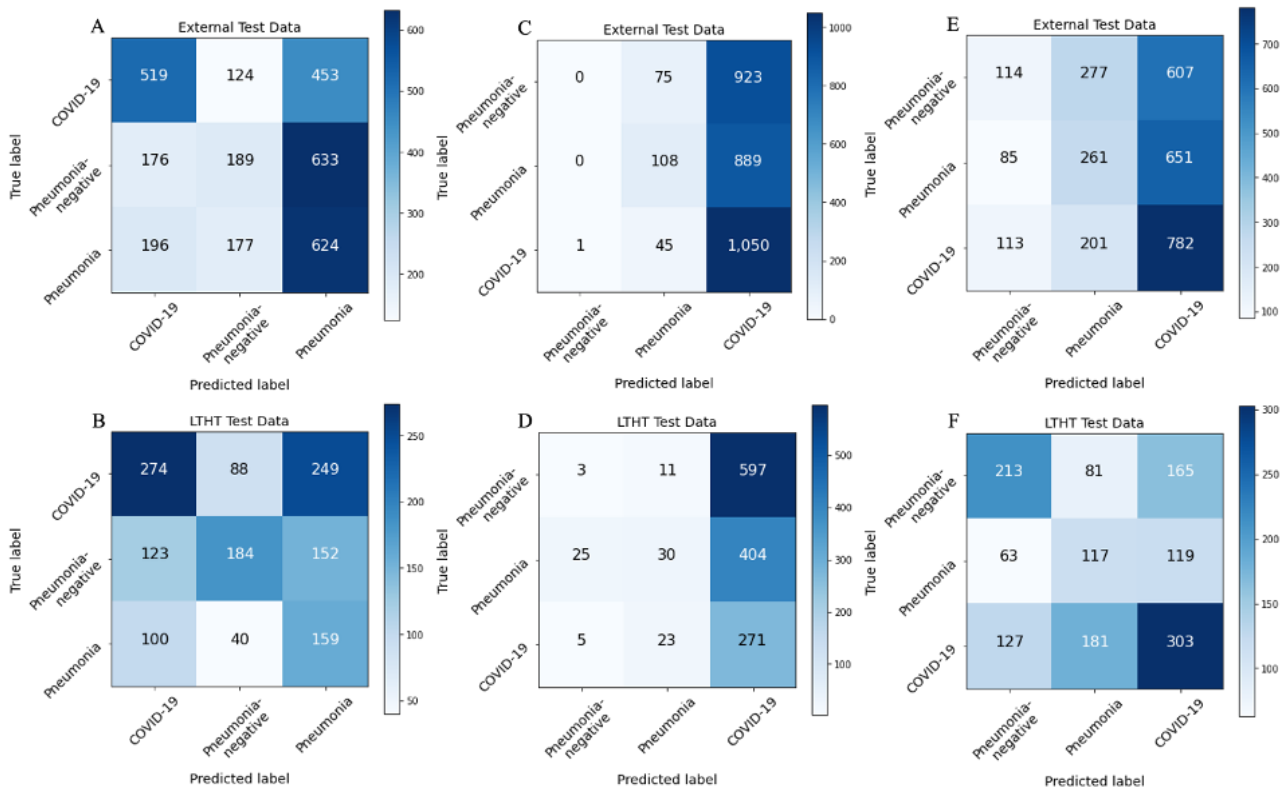

*Figure 4: Confusion matrices of prediction of external data for (A) DarkCovidNet, (C) CoroNet and (E) COVIDNet. Confusion matrices of prediction of LTHT data (B) DarkCovidNet, (D) CoroNet and (F) COVIDNet*

Model performance metrics across the different test sets are displayed in Tables 1-3, these are averaged over all classes. These show a clear decrease when comparing performance on the COVIDx test set with performance on the external and LTHT set data. Significant misclassification of all classes is identified (Figure 4) when testing on the external test data with the LTHT data demonstrating an inability to differentiate clinically-realistic data classes.

A clinical review of 500 grad-CAM saliency maps generated by prediction on COVIDx test data showed a trend of significance in clinically irrelevant features. This commonly included a focus on bony-structures and soft tissues instead of diffuse bilateral opacification of the lung fields that are typical of COVID-19 infection (Fig 5).

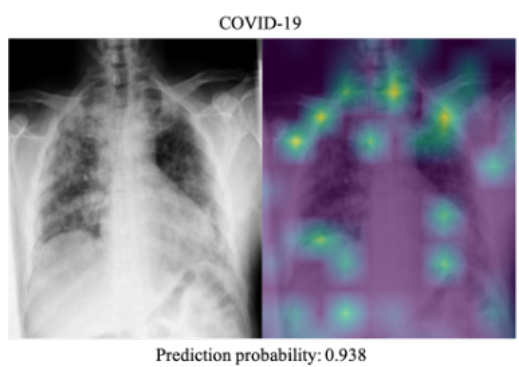

*Figure 5: Grad-CAM saliency map generated by DarkCovidNet COVIDx test data predictions. This is a correct prediction of COVID-19, with a prediction probability of 0.938*

**Discussion**

The issues identified within the COVIDx dataset raise a number significant concerns. While the application of less precise class labels may be suitable for the segmentation task the RSNA pneumonia detection challenge is designed for, it is not appropriate for diagnostic tasks where disease specificity is key. The COVIDx dataset expands the definition of pneumonia to include other confounding pathologies. This results in a dataset comprised of clinically inappropriate diagnostic labels that artificially make the classes of interest more distinct, thereby simplifying the task. Separation of COVID-19 from non-COVID pneumonia is a more challenging task than the separation of pneumonia from other lung pathologies. However, testing on COVIDx shows that the models are able to reliably separate COVID-19 from pneumonia, nearly as well as they are able to distinguish the normal class from the pneumonia class. This is surprising, particularly as any pneumonia-like chest X-rays have been excluded from the 'normal' class of the COVIDx dataset. With this in mind, the expansion of the pneumonia class also raises concerns as to whether COVID-19 cases are truly being distinguished from pneumonia cases as reported, or if they are being separated from the alternative classes included in the pneumonia class of COVIDx.

Testing on external and hospital data shows the models trained on COVIDx are not robust. A small decrease in performance can be expected when testing on external datasets due to domain shift, however, with models achieving no more than 45% prediction accuracy (Table 1-3) this drop is too severe to be entirely a result of this. This finding supports the idea that model performance may be falsely inflated when evaluated on the artificially simplistic COVIDx test data. The external test set is curated to reflect clinically realistic problems. The non-pneumonia class includes alternative pathologies and the pneumonia-positive class includes exclusively non-COVID pneumonia cases, alongside any number of comorbidities. While this brings dataset closer to the clinical problem, it also increases complexity of the problem and makes classification more challenging. Performance metrics collected from testing the models on the external test set (Tables 1-3) indicate that models trained on COVIDx are unable to generalise to clinically-realistic data. This is supported by the poor performance of all models on the LTHT dataset, which is best exemplified by the CoroNet model, which shows a decrease in prediction accuracy of 66% when testing on clinical data versus COVIDx data (Table 1).

The heterogeneity of the COVIDx data may also be contributing to the problem. While clinical features are consistent across data sources, non-clinical image features, such as, image size, image acquisition and image markers vary. When relying exclusively on one data source to supply a single class of images the model is made vulnerable to bias and can learn to rely on the non-clinical features unique to the data source over the clinically significant features of the disease. Studies that do not employ exact replicas of the COVIDx dataset still take the approach of forming heterogenous dataset by combining chest X-ray repositories from the pre-COVID era with COVID-19 data sources. This suggests that our findings are likely to apply to the whole domain and perhaps to other problem domains that rely on similar approaches for data collation

A critical issue in the use of open-source data is the lack of patient information. Without access to the demographic or clinical data associated with the chest X-ray data it is impossible to identify or account for confounding factors, such as, age. Manual assessment of the COVIDx dataset shows this factor is likely impacting the models assessed in this study. The inclusion of paediatric images almost exclusively within the non-COVID pneumonia component of COVIDx likely results in significant confounding, with the non-COVID class being identified not based on true clinical features, but rather paediatric chest X-ray features. This issue is likely to be further compounded by the increased probability that older age groups will require hospitalisation and radiographical testing if they contract COVID-19. Confounding variables can be controlled for through data augmentation techniques but without full access to the patient age information this becomes impossible. Similarly, without this information in-depth analysis of the impact of bias and confounding on model performance is not feasible. Clinical review of grad-CAM saliency maps confirm that, despite strong prediction performance on COVIDx test data, models rely on clinically irrelevant features with models often relying on bias and confounding originating in the COVIDx dataset.

A lack of available hospital data combined with inadequate model evaluation across the problem domain has allowed the use of open-source data to mislead the research community. Continued publication of inflated model performance metrics risks damaging the trustworthiness of AI research in medical diagnostics, particularly where the disease is of great public interest. The quality of research in this domain must improve to prevent this from happening, this must start with the data.

## Conclusion

Analysis of the COVIDx dataset showed misuse of labels as well as high risk of bias and confounding. Poor performance of deep learning models trained on the COVIDx dataset on both publicly available external test data and LTHT data demonstrates that the exceptional performance reported widely across the problem domain is inflated, that model performance results are misrepresented, and that models do not generalise well to clinically-realistic data. This demonstrates the need for greater access to clinical data, representative of the clinical problem, to facilitate thorough model evaluations and solve real clinical problems. Bridging this gap is crucial in the development of robust AI tools for medical diagnosis.

## Acknowledgements

This work uses data provided by patients and collected by the NHS as part of their care. This study was supported by AWS cloud computing credits awarded through the Diagnostic Development Initiative.

## Address for correspondence

Rachael Harkness: scrha@leeds.ac.uk